\documentclass[conference]{IEEEtran}
\usepackage[letterpaper, top=0.75in, bottom=0.95in, left=0.63in, right=0.63in]{geometry}
\usepackage[nolist]{acronym}
\usepackage{float}
\usepackage{fontawesome}
\usepackage{amsmath}
\usepackage{tikz}

\usepackage{graphicx}
\usepackage{xcolor}     % Para cores (mais moderno que o color)
\usepackage{tabularray}  % OBRIGATÓRIO para o ambiente tblr funcionar
\let\labelindent\relax
\usepackage{enumitem}
\usepackage[T1]{fontenc}
\usepackage{url}
\usepackage{booktabs}
\usepackage{multirow}
\usetikzlibrary{arrows.meta,positioning,calc,shadows}
\usepackage{fvextra}
\newfloat{listing}{htbp}{lol}
\floatname{listing}{Listing}
\input{iod_pygments_style.tex}

\usepackage{algorithm}
\usepackage{algorithmic}
\usepackage[caption=false,font=footnotesize]{subfig}

\definecolor{HeaderSoftBlue}{HTML}{D9EAF7}
\definecolor{OurSoftGray}{HTML}{F2F2F2}

\ifCLASSINFOpdf
\else
\fi
\begin{document}
%Numbering
\acrodef{3GPP}{3rd Generation Partnership Project}
\acrodef{5G}{5th Generation Mobile Network}
\acrodef{6G}{6th Generation Mobile Network}
%-----A-----
\acrodef{AI}{Artificial Intelligence}
\acrodef{AI4Net}{\ac{AI} for Networking}
\acrodef{MDP}{Markov Decision Process}
\acrodef{AIDER}{Aerial Image Dataset for Emergency Response}
\acrodef{AMF}{Access and Mobility Management Function}
\acrodef{AIaaS}{Artificial Intelligence-as-a-Service}
\acrodef{AC}{Actor-Critic}
\acrodef{AIDER}{Aerial Image Database for Emergency Response applications}
\acrodef{API}{Application Programming Interface}
\acrodef{BESS}{Berkeley Extensible Software Switch}
%-----B-----
\acrodef{B5G}{Beyond Fifth Generation}
\acrodef{BPF}{Berkeley Packet Filter}
%-----C-----
\acrodef{CBR}{Constant Bit Rate}
\acrodef{CSV}{Comma-Separated Values}
\acrodef{CPU}{Central Processing Unit}
\acrodef{CNN}{Convolutional Neural Network}
\acrodef{CNNs}{Convolutional Neural Networks}
\acrodef{C-V2X}{Cellular Vehicle to-Everything}
%-----D-----
\acrodef{DoS}{Denial of Service}
\acrodef{DDQL}{Double Deep
 Q-learning}
\acrodef{DDoS}{Distributed Denial of Service}
\acrodef{DDPG}{Deep Deterministic Policy Gradient}
\acrodef{DNN}{Deep Neural Network}
\acrodef{DRL}{Deep Reinforcement Learning}
\acrodef{DQN}{Deep Q-Network}
\acrodef{DT}{Decision Tree}
\acrodef{DDQN}{Double Deep Q-Network}

%-----E-----
\acrodef{ETSI}{European Telecommunications Standards Institute}
\acrodef{eNWDAF}{Evolved Network Data Analytics Function}
\acrodef{eBPF}{Extended Berkeley Packet Filter}
\acrodef{ECDF}{Empirical Cumulative Distribution Function}
\acrodef{ECDFs}{Empirical Cumulative Distribution Functions}
%-----F-----
\acrodef{FIBRE}{Future Internet Brazilian Environment for Experimentation}
\acrodef{FL}{Federated Learning}
\acrodef{FEL}{Federated Ensemble Learning}
%-----G-----
\acrodef{GNN}{Graph Neural Networks}
\acrodef{GPU}{Graphics Processing Unit}
\acrodef{GTP}{GPRS Tunnelling Protocol}
\acrodef{GTP-U}{GPRS Tunnelling Protocol User Plane}
\acrodef{GA}{Genetic Algorithm}
%-----H-----
\acrodef{HTM}{Hierarchical Temporal Memory}

%-----I-----
\acrodef{IAM}{Identity And Access Management}
\acrodef{ICMP}{Internet Control Message Protocol}
\acrodef{IID}{Independent and Identically Distributed}
\acrodef{IoE}{Internet of Everything}
\acrodef{IoT}{Internet of Things}
\acrodef{ITU}{International Telecommunication Union}
\acrodef{IQR}{Interquartile Range}
\acrodef{I/O}{Input/Output}
\acrodef{IP}{Internet Protocol}
\acrodef{IoD}{Inference-on-Demand}
\acrodef{IPv6}{Internet Protocol version 6}
%-----J-----
%-----K-----
\acrodef{KNN}{K-Nearest Neighbors}
\acrodef{KPI}{Key Performance Indicator}
\acrodef{KPIs}{Key Performance Indicators}
%-----L-----
\acrodef{LSTM}{Long Short-Term Memory}
\acrodef{LOWESS}{Locally Weighted Scatterplot Smoothing}
\acrodef{LR}{Learning Rate}
\acrodef{LLM}{Large Language Model}
%-----M-----
\acrodef{MAE}{Mean Absolute Error}
\acrodef{MAD}{Median Absolute Deviation}
\acrodef{ML}{Machine Learning}
\acrodef{MLaaS}{Machine Learning as a Service}
\acrodef{MOS}{Mean Opinion Score}
\acrodef{MAPE}{Mean Absolute Percentage Error}
\acrodef{MSE}{Mean Squared Error}
\acrodef{MEC}{Multi-access Edge Computing}
\acrodef{mMTC}{Massive Machine Type Communications}
\acrodef{MFA}{Multi-factor Authentication}
\acrodef{MLP}{Multi-Layer Perceptron}
\acrodef{MADRL}{Multi-Agent Deep Reinforcement Learning}
\acrodef{MAB}{Multi-Armed Bandit}
\acrodef{MILP}{Mixed Integer Linear Programming}
\acrodef{MQTT}{Message Queuing Telemetry Transport}
%-----N-----
\acrodef{NWDAF}{Network Data Analytics Function}
\acrodef{Net4AI}{Networking for \ac{AI}}
\acrodef{NS}{Network Slicing}
\acrodef{NFV}{Network Function Virtualization}
\acrodef{NEB}{Net Energy Benefit}
%-----O-----
\acrodef{OSM}{Open Source MANO}
%-----P-----
\acrodef{PCA}{Principal Component Analysis}
\acrodef{PC-FedAvg}{Personalized Conditional Federated Averaging}
\acrodef{PoC}{Proof of Concept}
\acrodef{PPO}{Proximal Policy Optimization}
\acrodef{POMDP}{Partially Observable Markov decision process}
\acrodef{PCAP}{Packet Capture}
\acrodef{PSO}{Particle Swarm Optimization}
\acrodef{PDU}{Power Distribution Unit}
\acrodef{PFCP}{Packet Forwarding Control Protocol}
\acrodef{PDR}{Packet Detection Rule}
%-----Q-----
\acrodef{QoE}{Quality of experience}
\acrodef{QoS}{Quality of Service}
\acrodef{QFI}{QoS Flow Identifier}
\acrodef{QFIs}{QoS Flow Identifiers}
%-----R-----
\acrodef{RAM}{Random Access Memory}
\acrodef{RF}{Random Forest}
\acrodef{RL}{Reinforcement Learning}
\acrodef{RMSE}{Root Mean Square Error}
\acrodef{RNN}{Recurrent Neural Network}
\acrodef{RTT}{Round-Trip Time}
\acrodef{RAN}{Radio Access Network}
\acrodef{RTP}{Real-time Transport Protocol}
\acrodef{RAPL}{Running Average Power Limit}
%-----S-----
\acrodef{SDN}{Software-Defined Networking}
\acrodef{SFI2}{Slicing Future Internet Infrastructures}
\acrodef{SLA}{Service-Level Agreement}
\acrodef{SON}{Self-Organizing Network}
\acrodef{SMF}{Session Management Function}
\acrodef{S-NSSAI}{Single Network Slice Selection Assistance Information}
\acrodef{SVM}{Support Vector Machine}
\acrodef{SOPS}{Service-Aware Optimal
 Path Selection}
 \acrodef{SAFE}{Scalable Asynchronous Federated Ensembling}
 \acrodef{SRv6}{Segment Routing over IPv6}
 \acrodef{SRH}{Segment Routing Extension Header}
%-----T-----
\acrodef{TQFL}{Trust Deep Q-learning Federated Learning}
\acrodef{TEID}{Tunnel Endpoint Identifier}
\acrodef{TEIDs}{Tunnel Endpoint Identifiers}
\acrodef{TPE}{Tree-Structured Parzen Estimator}
\acrodef{TLV}{Type-Length-Value}
%-----U-----
\acrodef{UE}{User Equipment}
\acrodef{UEs}{User Equipments}
\acrodef{UPF}{User Plane Function}
\acrodef{UPFs}{User Plane Functions}
\acrodef{PDU}{Packet Data Unit}
\acrodef{URLLC}{Ultra-Reliable and Low Latency Communications}
\acrodef{UAV}{Unmanned Aerial Vehicle}
\acrodef{UAVs}{Unmanned Aerial Vehicles}
\acrodef{UDP}{User Datagram Protocol}
%-----V-----
\acrodef{VoD}{Video on Demand}
\acrodef{VR}{Virtual Reality}
\acrodef{AR}{Augmented Reality}
\acrodef{V2V}{Vehicle-to-Vechile}
\acrodef{V2X}{Vehicle-to-Everything}
\acrodef{VNF}{Virtual Network Function}
\acrodef{VNFs}{Virtual Network Functions}
\acrodef{VPN}{Virtual Private Network}
%-----W-----
%-----X-----
\acrodef{XDP}{eXpress Data Path}
%-----Y-----
%-----Z-----
%\title{Forecasting User Experience on Mobile Networks through Infrastructure Monitoring in container-based environments}
%\title{Forecasting User Experience in Mobile Networks through Generic Monitoring-Driven Approaches}
%\title{Leveraging Infrastructure Monitoring for User Experience Forecasting in container-based \\ Mobile Network Core}
%\title{Leveraging Infrastructure Monitoring for User Experience Forecasting in Container-based 5G Core}

%\title{Using Post-Quantum Cryptography to Enhance Resilience Against Cyberattacks in 5G and Future Networks}

%\title{Energy Aware Agentic Orchestration for Geographically Distributed Service Deployment}
\title{Exposing AI Inference as a Telco Network API over the 5G User Plane}

\author{\IEEEauthorblockN{Rodrigo Moreira\textsuperscript{1}, Larissa Ferreira {Rodrigues Moreira}\textsuperscript{1}, Joberto S. B. Martins\textsuperscript{2},\\ Tereza C. Carvalho\textsuperscript{3}, Flávio de Oliveira Silva\textsuperscript{4}}
\IEEEauthorblockA{
\textsuperscript{1}Federal University of Viçosa (UFV), Minas Gerais, Brazil\\
\textsuperscript{2}University of Salvador (UNIFACS), Salvador, Brazil\\
\textsuperscript{3}University of São Paulo (USP), São Paulo, Brazil\\
\textsuperscript{4}University of Minho (UMinho), Braga, Portugal\\
Emails: rodrigo@ufv.br, larissa.f.rodrigues@ufv.br, \\joberto@ieee.org, terezacarvalho@usp.br, flavio@di.uminho.pt}
}


%Telephone: (800) 555--1212, Fax: (888) 555--1212}


\maketitle

\begingroup
\renewcommand\thefootnote{}
\footnotetext{\scriptsize
© 2026 IEEE. Personal use of this material is permitted. Permission from IEEE must be obtained for all other uses, in any current or future media, including reprinting/republishing this material for advertising or promotional purposes, creating new collective works, for resale or redistribution to servers or lists, or reuse of any copyrighted component of this work in other works.

Accepted author manuscript for publication in the Proceedings of the 2026 IEEE Future Networks World Forum (FNWF 2026), Lisbon, Portugal, October 26--28, 2026.
}
\addtocounter{footnote}{-1}
\endgroup

\begin{abstract}
Supporting edge Artificial Intelligence (AI) inference as a network service is challenging because current mobile systems can transport tensors but do not expose or understand the inference context that gives those tensors operational meaning. Prior studies have addressed this gap through dynamic partitioning, collaborative inference, and split pipelines; however, these solutions remain application-bound, network-blind, or context-free. This work proposes Inference on Demand (IoD), an inference-on-demand abstraction that exposes split inference through a CAMARA-like Application Program Interface (API) and carries a split-aware tensor context in-band across the 5G user plane path. We evaluate the IoD on a programmable user plane function (UPF)-based testbed using a partitionable visual AI workload, showing that adaptive split selection can operate as a network-facing service while avoiding the rigid cost profile of static configurations. Including our findings, split inference becomes more than offloading when the service intent, runtime partitioning, and tensor transport are made visible to the mobile data plane. This opens a path toward network-native AI services, in which programmable telco infrastructure coordinates computation, context, and connectivity for user-facing applications.
\end{abstract}

\IEEEpeerreviewmaketitle

\section{Introduction}\label{sec:introduction}

Modern applications increasingly require networks to perform more than just packet delivery ~\cite{Tang2026}. Real-time \ac{AI}, immersive sensing, and safety-critical visual analytics depend on latency, compute placement, and context that are not fully captured by best-effort connectivity alone~\cite{Tang2026, Niu2026}. This gap has motivated open telco infrastructure initiatives, such as the GSMA Open Gateway and CAMARA, which expose network capabilities through common \acp{API} and make operator infrastructure programmable for developers~\cite{gsma_opengateway_alt,camaraproject, Ordonez-Lucena2022}. At the same time, the growth of \ac{5G} traffic and edge-native workloads has made softwarized mobile cores, programmable user planes, and \ac{AI}-assisted service control practical enablers for turning connectivity into a common programmable substrate for user-facing, context-aware applications~\cite{Niu2026}.

However, this broader role for the network is not available for edge inference today~\cite{Lin2025, Liu2026}. The problem is that split inference is still mostly treated as an application-offloading technique, rather than as a programmable network service~\cite{Lessa2025, Kim2026, rodrigo2026}. Existing studies have explored dynamic partitioning, but often outside the mobile user plane; collaborative inference, but without a telco service abstraction; and split-computing pipelines, but without inference-specific in-band context~\cite{paper5, Zhang2026}. The network can carry tensors, but it does not yet understand the inference session to which they belong. We propose \ac{IoD}, which turns split inference into a network-facing service primitive: the application requests an inference envelope, the \ac{UE} adapts the split point at runtime, and the user plane carries the inference context with the tensor fragments, making the mobile network an active substrate for adaptive \ac{AI} execution rather than a passive transport path~\cite{Niu2026}.

To address this gap, we designed and evaluated \ac{IoD}, a CAMARA-inspired service interface and user-plane mechanism that exposes adaptive split inference as a programmable network service. Rather than treating the mobile network as a passive path for tensor fragments, the \ac{IoD} lets the application negotiate an inference envelope, enables the \ac{UE} to adapt the split point at runtime, and binds activation traffic to inference-session context through in-band metadata carried in the \ac{GTP-U}/\ac{SRv6} packet stack. The contributions are threefold: (i) an application-facing abstraction for requesting and controlling split-inference sessions; (ii) a data-plane mechanism that transports inference-specific context together with activation fragments; (iii) and an experimental validation over a \ac{5G} user plane path showing how this service primitive can support adaptive \ac{AI} execution inside the mobile network.

The remainder of this work is organized as follows. Section~\ref{sec:related_work} reviews related work. Section~\ref{sec:method} presents the proposed method. Section~\ref{subsec:experiental_testbed} describes the experimental testbed. Section~\ref{sec:results} discusses the results. Section~\ref{sec:concluding_remarks} concludes the paper.

\section{Related Work}\label{sec:related_work}

Recent work has mainly advanced split inference as an optimization problem over computation, communication, and service constraints. Several studies adapt the split point or layer placement to improve latency, energy, throughput, or Service-Level Agreement (SLA) compliance in Mobile Edge Computing (MEC) systems: \cite{paper1} and \cite{paper3} target runtime partitioning and placement of Large Foundation Models (LFMs), while \cite{paper4}, \cite{paper6}, \cite{paper10}, \cite{paper11}, \cite{paper12}, \cite{paper13}, and \cite{paper14} optimize collaborative Deep Neural Network (DNN) inference across mobile devices, edge servers, or Device-to-Device (D2D) settings. Related learning-oriented works, such as \cite{paper2}, \cite{paper7}, and \cite{paper9}, extend this direction toward Extended Reality (XR), satellite edge computing, and vehicular 6G systems, showing that adaptive partitioning is increasingly treated as a runtime control problem rather than a static model-design choice.

A second line of work moves closer to the network infrastructure itself. \cite{paper5} proposes 5G architectural enhancements for split-computing service pipelines using Computing-Aware Traffic Steering (CATS), Segment Routing over IPv6 (SRv6), and Mobile User Plane (MUP), whereas \cite{paper8} studies functional split of in-network Deep Learning (DL) across enhanced User Plane Functions (UPFs). These works show that split Artificial Intelligence (AI) is no longer only an application-layer concern. However, they do not expose inference as a CAMARA-style telco Network Application Programming Interface (API), nor do they carry inference-specific split metadata in-band with the user-plane tensor traffic.

\begin{table}[htbp]
\centering
\scriptsize
\setlength{\tabcolsep}{4.0pt}
\renewcommand{\arraystretch}{1.14}
\newcommand{\full}{\faCircle}
\newcommand{\partly}{\faAdjust}
\newcommand{\emptyc}{\faCircleO}
\caption{Short state-of-the-art Survey}
\label{tab:sota_short_survey}
\resizebox{\columnwidth}{!}{%
\begin{tabular}{lcccccc}
\hline
\textbf{Approach} & \textbf{\begin{tabular}[c]{@{}c@{}}Dynamic \\ Split\end{tabular}} & \textbf{\begin{tabular}[c]{@{}c@{}}QoS \\ Control\end{tabular}} & \textbf{\begin{tabular}[c]{@{}c@{}}Tensor \\ Transport\end{tabular}} & \textbf{\begin{tabular}[c]{@{}c@{}}User \\ Plane\end{tabular}} & \textbf{\begin{tabular}[c]{@{}c@{}}In-Band \\ Metadata\end{tabular}} & \textbf{\begin{tabular}[c]{@{}c@{}}CAMARA \\ API\end{tabular}} \\ \hline
Djuhera et al.~\cite{paper1}       & \full                                                             & \full                                                           & \partly                                                              & \partly                                                        & \emptyc                                                              & \emptyc                                                        \\
Shah et al.~\cite{paper2}        & \full                                                             & \full                                                           & \full                                                                & \partly                                                        & \emptyc                                                              & \emptyc                                                        \\
Koch et al.~\cite{paper3}        & \full                                                             & \full                                                           & \partly                                                              & \partly                                                        & \emptyc                                                              & \emptyc                                                        \\
Zhang et al.~\cite{paper4}       & \full                                                             & \full                                                           & \full                                                                & \emptyc                                                        & \emptyc                                                              & \emptyc                                                        \\
Tran et al.~\cite{paper5}        & \emptyc                                                           & \partly                                                         & \partly                                                              & \full                                                          & \partly                                                              & \emptyc                                                        \\
Lim et al.~\cite{paper6}        & \full                                                             & \full                                                           & \full                                                                & \emptyc                                                        & \emptyc                                                              & \emptyc                                                        \\
Fang et al.~\cite{paper7}        & \partly                                                           & \full                                                           & \full                                                                & \emptyc                                                        & \emptyc                                                              & \emptyc                                                        \\
He et al.~\cite{paper8}        & \partly                                                           & \partly                                                         & \full                                                                & \full                                                          & \emptyc                                                              & \emptyc                                                        \\
Rasheed \& Mostafa~\cite{paper9}        & \full                                                             & \full                                                           & \full                                                                & \emptyc                                                        & \emptyc                                                              & \emptyc                                                        \\
Lim \& Kim~\cite{paper10}       & \full                                                             & \full                                                           & \full                                                                & \emptyc                                                        & \emptyc                                                              & \emptyc                                                        \\
Fang et al.~\cite{paper11}       & \partly                                                           & \full                                                           & \full                                                                & \emptyc                                                        & \emptyc                                                              & \emptyc                                                        \\
Tan et al.~\cite{paper12}       & \full                                                             & \full                                                           & \full                                                                & \emptyc                                                        & \emptyc                                                              & \emptyc                                                        \\
Tan et al.~\cite{paper13}       & \full                                                             & \full                                                           & \full                                                                & \emptyc                                                        & \emptyc                                                              & \emptyc                                                        \\
Guo et al.~\cite{paper14}       & \full                                                             & \partly                                                         & \full                                                                & \emptyc                                                        & \emptyc                                                              & \emptyc                                                        \\ \hline
\textbf{Our}      & \full                                                             & \full                                                           & \full                                                                & \full                                                          & \full                                                                & \full                                                          \\ \hline
\end{tabular}%
}
\end{table}

Table~\ref{tab:sota_short_survey} summarizes this landscape through six dimensions: Dynamic Split means runtime partition adaptation; QoS Control means service objective enforcement; Tensor Transport means activation payload transfer; User Plane means telco datapath integration; In-Band Metadata means packet-carried inference context; CAMARA API means programmable service exposure.

\section{Proposed Method}\label{sec:method}

\begin{figure}[htbp]
\centering
\includegraphics[width=1\columnwidth]{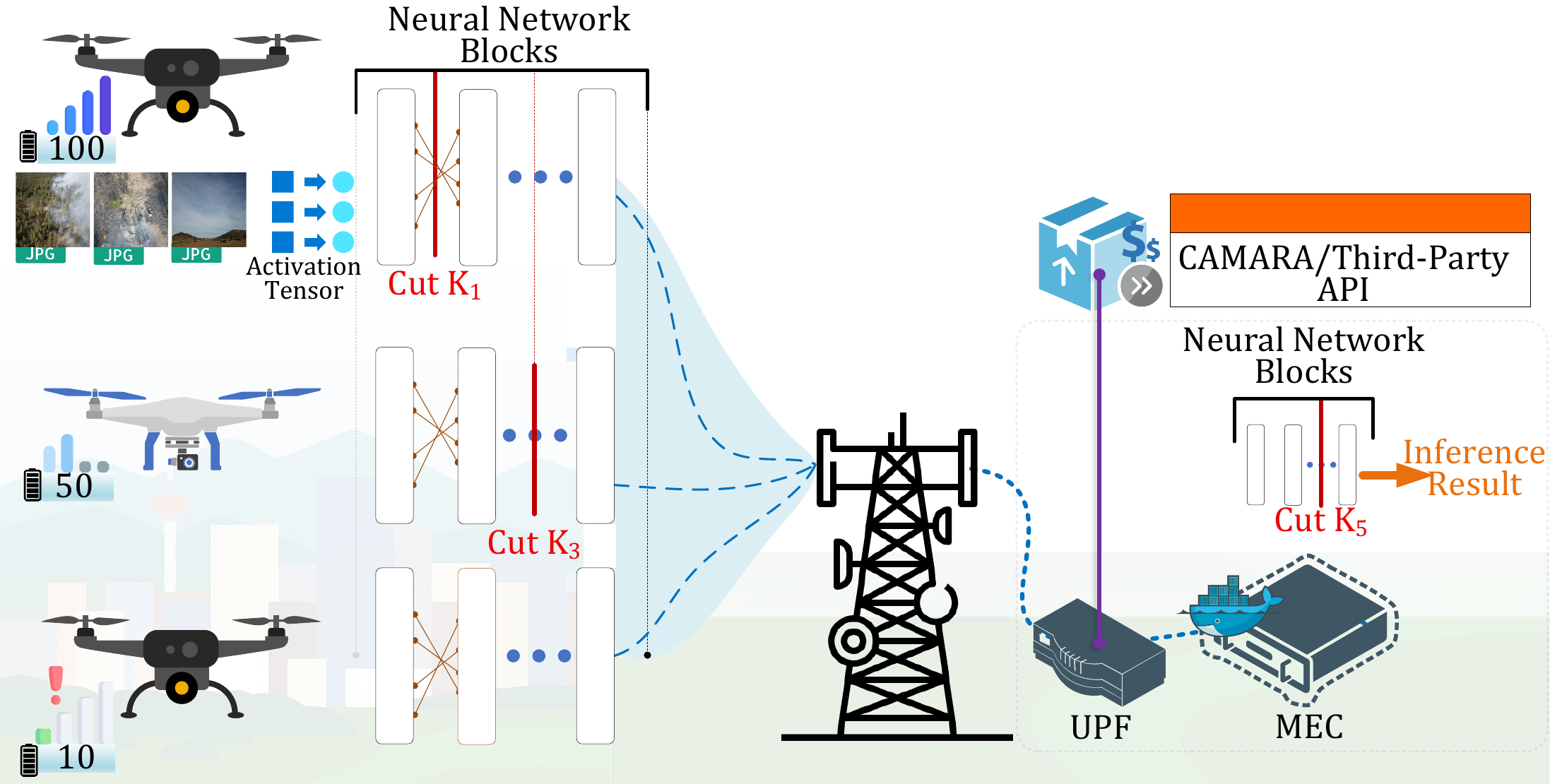}
\caption{Inference-on-Demand problem setting.}
\label{fig:method}
\end{figure}

Figure~\ref{fig:method} shows the core idea of our approach: if telco networks are to host new services, their capabilities must be exposed as programmable network \acp{API}, not only as connectivity pipes. \ac{IoD} follows this direction by exposing edge inference as a CAMARA-style~\cite{camaraproject} service. The split point $k$ is the service knob: smaller $k$ protects the \ac{UE} but pushes larger tensors into the user plane, whereas larger $k$ spends more local computing resources but sends fewer bytes to the \ac{MEC}. For catastrophic scenarios, where device and network conditions change over time, fixing $k$ in advance is excessively rigid. \ac{IoD} keeps the model partitionable and lets the service adapt where inference crosses from the \ac{UE} to the network.

Let a neural model be represented as an ordered sequence of layers 
$\mathcal{M}={L_1,L_2,\ldots,L_N}$. A split point $k$ partitions the model into a UE-side head $\mathcal{M}*{\mathrm{UE}}^{k}={L_1,\ldots,L_k}$ and an edge-side tail $\mathcal{M}*{\mathrm{MEC}}^{k}={L_{k+1},\ldots,L_N}$. Given an input frame $x_t$, the UE computes an intermediate activation $z_t=\mathcal{M}_{\mathrm{UE}}^{k}(x_t)$ and sends this activation, rather than the original frame, to the MEC. The final prediction is then computed as $\hat{y}*t=\mathcal{M}*{\mathrm{MEC}}^{k}(z_t)$. Smaller values of $k$ offload earlier layers and reduce UE-side computation, but they generate larger intermediate tensors. Larger values of $k$ execute more layers locally and reduce the transmitted activation size, but increase UE-side processing. IoD exploits this trade-off by selecting $k$ online from an admissible set $\mathcal{K}$, instead of using a single static split throughout the service session.

\begin{algorithm}[htbp]
\caption{Runtime workflow of \ac{IoD} split inference.}
\label{alg:iod_runtime}
\begin{algorithmic}[1]
\REQUIRE Service intent $I=(\mathcal{K},\tau_{\mathrm{tx}},\tau_{\mathrm{head}},q)$, frame $x_t$, \ac{TEID}
\STATE Observe \ac{UE} and path state $s_t$ from local execution and recent transmission
\STATE Select $k_t \leftarrow \pi_I(s_t)$, with $k_t \in \mathcal{K}$
\STATE Compute $z_t \leftarrow \mathcal{M}^{k_t}_{\mathrm{UE}}(x_t)$
\STATE Encode and fragment $z_t$ according to quantization mode $q$
\FOR{each fragment $p_j$}
\STATE Write $(k_t,q,j,m,\mathrm{shape}(z_t))$ into the split metadata \ac{TLV}
\STATE Encapsulate $p_j$ as GTP-U(TEID)/IPv6/SRH/TLV/payload and send on N3
\ENDFOR
\STATE UPF matches the TEID, decapsulates GTP-U, parses the split \ac{TLV}, and forwards to N6/MEC
\STATE \ac{MEC} reassembles $z_t$ and computes $\hat{y}_t \leftarrow \mathcal{M}^{k_t}_{\mathrm{MEC}}(z_t)$
\end{algorithmic}
\end{algorithm}

Figure~\ref{alg:iod_runtime} summarizes the runtime workflow. The service intent defines the admissible split range, thresholds, and quantization mode; the \ac{UE} then selects $k_t=\pi_I(s_t)$ from its local execution and recent transmission state. Unlike a fixed split, \ac{IoD} does not replace the model or reconfigure the \ac{MEC} service at each frame: it only changes where the current inference is partitioned. The \ac{UE} transmits encoded activation tensors rather than raw frames, with compact metadata for split identification, shape, type, and reassembly. This makes the abstraction relevant to privacy-sensitive visual applications, where reducing the exposure of raw images in the user plane is part of the service requirement.

In our approach we build a CAMARA-like~\cite{camaraproject} \ac{IoD} \ac{API}. The \ac{API} does not micromanage every frame; it opens the telco infrastructure to a new class of services by negotiating an inference envelope. As shown in Listing~\ref{lst:iod_api}, the application requests a model, an admissible range of split points, and a quantization policy. The network grants a service session with a \ac{MEC} target and a data-plane metadata schema. The concrete split point $k_t$ is then selected at runtime by the \ac{UE}, inside the admitted range, and carried in-band with the activation tensor.

% Code pre-highlighted offline using Pygments JSON lexer; no minted or shell escape.
\begin{listing}[htbp]
\begin{Verbatim}[commandchars=\\\{\},fontsize=\scriptsize,numbers=left,firstnumber=1,numbersep=8pt,frame=topline,framesep=2mm,breaklines=true,breakanywhere=true,baselinestretch=0.95]
\PY{c+c1}{// \PYZhy{}\PYZhy{}\PYZhy{} 1) Service Request \PYZhy{}\PYZhy{}\PYZhy{}}
\PY{p}{\PYZob{}}
\PY{+w}{  }\PY{n+nt}{\PYZdq{}applicationId\PYZdq{}}\PY{p}{:}\PY{+w}{ }\PY{l+s+s2}{\PYZdq{}aerial\PYZhy{}fire\PYZhy{}detection\PYZdq{}}\PY{p}{,}
\PY{+w}{  }\PY{n+nt}{\PYZdq{}serviceType\PYZdq{}}\PY{p}{:}\PY{+w}{ }\PY{l+s+s2}{\PYZdq{}inference\PYZhy{}on\PYZhy{}demand\PYZdq{}}\PY{p}{,}
\PY{+w}{  }\PY{n+nt}{\PYZdq{}model\PYZdq{}}\PY{p}{:}\PY{+w}{ }\PY{p}{\PYZob{}}\PY{n+nt}{\PYZdq{}name\PYZdq{}}\PY{p}{:}\PY{+w}{ }\PY{l+s+s2}{\PYZdq{}flame2\PYZhy{}yolo\PYZhy{}like\PYZdq{}}\PY{p}{,}\PY{+w}{ }\PY{n+nt}{\PYZdq{}task\PYZdq{}}\PY{p}{:}\PY{+w}{ }\PY{l+s+s2}{\PYZdq{}fire\PYZhy{}detection\PYZdq{}}\PY{p}{\PYZcb{},}
\PY{+w}{  }\PY{n+nt}{\PYZdq{}latencyObjectiveMs\PYZdq{}}\PY{p}{:}\PY{+w}{ }\PY{l+m+mi}{120}\PY{p}{,}
\PY{+w}{  }\PY{n+nt}{\PYZdq{}splitPolicy\PYZdq{}}\PY{p}{:}\PY{+w}{ }\PY{p}{\PYZob{}}
\PY{+w}{    }\PY{n+nt}{\PYZdq{}allowedK\PYZdq{}}\PY{p}{:}\PY{+w}{ }\PY{p}{[}\PY{l+m+mi}{2}\PY{p}{,}\PY{+w}{ }\PY{l+m+mi}{3}\PY{p}{,}\PY{+w}{ }\PY{l+m+mi}{4}\PY{p}{,}\PY{+w}{ }\PY{l+m+mi}{5}\PY{p}{],}
\PY{+w}{    }\PY{n+nt}{\PYZdq{}defaultK\PYZdq{}}\PY{p}{:}\PY{+w}{ }\PY{l+m+mi}{3}\PY{p}{,}
\PY{+w}{    }\PY{n+nt}{\PYZdq{}decisionScope\PYZdq{}}\PY{p}{:}\PY{+w}{ }\PY{l+s+s2}{\PYZdq{}ue\PYZhy{}runtime\PYZdq{}}
\PY{+w}{  }\PY{p}{\PYZcb{},}
\PY{+w}{  }\PY{n+nt}{\PYZdq{}quantization\PYZdq{}}\PY{p}{:}\PY{+w}{ }\PY{p}{\PYZob{}}\PY{n+nt}{\PYZdq{}allowed\PYZdq{}}\PY{p}{:}\PY{+w}{ }\PY{p}{[}\PY{l+s+s2}{\PYZdq{}fp16\PYZdq{}}\PY{p}{,}\PY{+w}{ }\PY{l+s+s2}{\PYZdq{}int8\PYZdq{}}\PY{p}{],}\PY{+w}{ }\PY{n+nt}{\PYZdq{}default\PYZdq{}}\PY{p}{:}\PY{+w}{ }\PY{l+s+s2}{\PYZdq{}fp16\PYZdq{}}\PY{p}{\PYZcb{},}
\PY{+w}{  }\PY{n+nt}{\PYZdq{}triggers\PYZdq{}}\PY{p}{:}\PY{+w}{ }\PY{p}{\PYZob{}}\PY{n+nt}{\PYZdq{}txDelayMs\PYZdq{}}\PY{p}{:}\PY{+w}{ }\PY{l+m+mi}{22}\PY{p}{,}\PY{+w}{ }\PY{n+nt}{\PYZdq{}headLatencyMs\PYZdq{}}\PY{p}{:}\PY{+w}{ }\PY{l+m+mi}{80}\PY{p}{\PYZcb{}}
\PY{p}{\PYZcb{}}
\end{Verbatim}
\vspace{-1.2em}
\begin{Verbatim}[commandchars=\\\{\},fontsize=\scriptsize,numbers=left,firstnumber=15,numbersep=8pt,frame=bottomline,framesep=2mm,breaklines=true,breakanywhere=true,baselinestretch=0.95]
\PY{c+c1}{// \PYZhy{}\PYZhy{}\PYZhy{} 2) Session Response \PYZhy{}\PYZhy{}\PYZhy{}}
\PY{p}{\PYZob{}}
\PY{+w}{  }\PY{n+nt}{\PYZdq{}sessionId\PYZdq{}}\PY{p}{:}\PY{+w}{ }\PY{l+s+s2}{\PYZdq{}iod\PYZhy{}7f31\PYZdq{}}\PY{p}{,}
\PY{+w}{  }\PY{n+nt}{\PYZdq{}status\PYZdq{}}\PY{p}{:}\PY{+w}{ }\PY{l+s+s2}{\PYZdq{}granted\PYZdq{}}\PY{p}{,}
\PY{+w}{  }\PY{n+nt}{\PYZdq{}mecTarget\PYZdq{}}\PY{p}{:}\PY{+w}{ }\PY{p}{\PYZob{}}
\PY{+w}{    }\PY{n+nt}{\PYZdq{}nodeId\PYZdq{}}\PY{p}{:}\PY{+w}{ }\PY{l+s+s2}{\PYZdq{}mec\PYZhy{}gpu\PYZhy{}01\PYZdq{}}\PY{p}{,}\PY{+w}{ }
\PY{+w}{    }\PY{n+nt}{\PYZdq{}n6Address\PYZdq{}}\PY{p}{:}\PY{+w}{ }\PY{l+s+s2}{\PYZdq{}192.168.90.2\PYZdq{}}\PY{p}{,}\PY{+w}{ }
\PY{+w}{    }\PY{n+nt}{\PYZdq{}port\PYZdq{}}\PY{p}{:}\PY{+w}{ }\PY{l+m+mi}{5001}
\PY{+w}{  }\PY{p}{\PYZcb{},}
\PY{+w}{  }\PY{n+nt}{\PYZdq{}admittedSplitRange\PYZdq{}}\PY{p}{:}\PY{+w}{ }\PY{p}{\PYZob{}}\PY{n+nt}{\PYZdq{}minK\PYZdq{}}\PY{p}{:}\PY{+w}{ }\PY{l+m+mi}{2}\PY{p}{,}\PY{+w}{ }\PY{n+nt}{\PYZdq{}maxK\PYZdq{}}\PY{p}{:}\PY{+w}{ }\PY{l+m+mi}{5}\PY{p}{,}\PY{+w}{ }\PY{n+nt}{\PYZdq{}fallbackK\PYZdq{}}\PY{p}{:}\PY{+w}{ }\PY{l+m+mi}{3}\PY{p}{\PYZcb{},}
\PY{+w}{  }\PY{n+nt}{\PYZdq{}dataPlaneMetadata\PYZdq{}}\PY{p}{:}\PY{+w}{ }\PY{p}{\PYZob{}}
\PY{+w}{    }\PY{n+nt}{\PYZdq{}carrier\PYZdq{}}\PY{p}{:}\PY{+w}{ }\PY{l+s+s2}{\PYZdq{}SRv6\PYZhy{}SRH\PYZhy{}TLV\PYZdq{}}\PY{p}{,}\PY{+w}{ }
\PY{+w}{    }\PY{n+nt}{\PYZdq{}tlvType\PYZdq{}}\PY{p}{:}\PY{+w}{ }\PY{l+m+mi}{124}\PY{p}{,}
\PY{+w}{    }\PY{n+nt}{\PYZdq{}fields\PYZdq{}}\PY{p}{:}\PY{+w}{ }\PY{p}{[}
\PY{+w}{      }\PY{l+s+s2}{\PYZdq{}k\PYZdq{}}\PY{p}{,}\PY{+w}{ }\PY{l+s+s2}{\PYZdq{}dtype\PYZdq{}}\PY{p}{,}\PY{+w}{ }\PY{l+s+s2}{\PYZdq{}fragmentOffset\PYZdq{}}\PY{p}{,}\PY{+w}{ }
\PY{+w}{      }\PY{l+s+s2}{\PYZdq{}fragmentTotal\PYZdq{}}\PY{p}{,}\PY{+w}{ }\PY{l+s+s2}{\PYZdq{}scale\PYZdq{}}\PY{p}{,}\PY{+w}{ }\PY{l+s+s2}{\PYZdq{}tensorShape\PYZdq{}}
\PY{+w}{    }\PY{p}{]}
\PY{+w}{  }\PY{p}{\PYZcb{}}
\PY{p}{\PYZcb{}}
\end{Verbatim}
\caption{CAMARA-like API interaction. The request (top) defines the envelope, while the response (bottom) binds it to a \ac{MEC} target and datapath metadata.}
\label{lst:iod_api}
\end{listing}

The \ac{UPF} is the user-plane anchor between the radio-side path and the \ac{MEC}-side service. In our architecture, the \ac{UE} sends activation fragments through the N3 interface using \ac{GTP}-U encapsulation toward the \ac{UPF}. Inside the \ac{GTP}-U payload, the packet carries an \ac{IPv6} and \ac{SRH} stack with a split-inference \ac{TLV} and the raw activation fragment. The relevant packet structure is according to Eq.~\ref{eq:protocol_stack}:

\begin{equation}
\scriptsize
\label{eq:protocol_stack}
\mathrm{Ethernet}/\allowbreak\mathrm{IP}/\allowbreak\mathrm{UDP}_{2152}/\allowbreak\mathrm{GTP\mbox{-}U}(\mathrm{TEID})/ \\
\allowbreak\mathrm{IPv6}/\allowbreak\mathrm{SRH}{+}\mathrm{TLV}/\allowbreak\mathrm{payload}.
\end{equation}

The \ac{TLV} carries the selected split point $k$, the tensor data type, fragment offset, total number of fragments, scaling information for quantized tensors, and tensor shape. Therefore, the split decision travels in-band with the data plane packet, while \ac{PFCP} remains responsible for session-level state and TEID installation.

In our implementation, the pipeline receives N3 packets, parses the outer headers, matches the session by \ac{TEID} through the \ac{PDR} state, and performs \ac{GTP}-U decapsulation. The split-inference handling is placed immediately after decapsulation, when the inner \ac{IPv6}/\ac{SRH} packet becomes visible to the data-plane pipeline. At this point, the \ac{UPF} can parse the split \ac{TLV}, account the traffic by \ac{TEID} and split
point, and forward the resulting payload toward the \ac{MEC} through the N6 side. The \ac{UPF} only exposes the inner packet after \ac{GTP}-U decapsulation, reads the in-band metadata, and keeps the inference traffic observable and forwardable inside the mobile user plane.

At the \ac{MEC}, \ac{TLV} metadata drives fragment reassembly, tail selection, and inference completion. The same partitionable model therefore admits several runtime operating points: later splits save user-plane bytes, earlier splits save \ac{UE} compute. \ac{IoD} ties these choices across three planes: the CAMARA-like \ac{API} defines the intent, the \ac{UE} selects $k_t$, and the \ac{UPF} carries and observes the activation traffic toward the \ac{MEC}.

%\begin{table}[t]
%\centering
%\caption{Split-inference TLV carried in-band with each activation fragment.}
%\label{tab:split_tlv}
%\footnotesize
%\begin{tabular}{lll}
%\toprule
%\textbf{Field} & \textbf{Meaning} & \textbf{Purpose} \\ % <-- Duas barras aqui
%\midrule
%$k$ & selected split point & selects the MEC-side model tail \\ 
%\texttt{dtype} & tensor numeric format & identifies FP32, FP16, or INT8 \\ 
%\texttt{frag\_off} & fragment index & orders the current fragment \\ % <-- Underscores escapados!
%\texttt{frag\_total} & total fragments & detects complete activation transfer \\ 
%\texttt{scale} & quantization scale & dequantizes INT8 activations \\ 
%\texttt{shape} & tensor dimensions & reconstructs the activation tensor \\ 
%\bottomrule
%\end{tabular}
%\end{table}

% ------------------------------------------------------------

\subsection{Experimental Testbed}\label{subsec:experiental_testbed}

\begin{figure}[htbp]
    \centering
    \includegraphics[width=1\columnwidth]{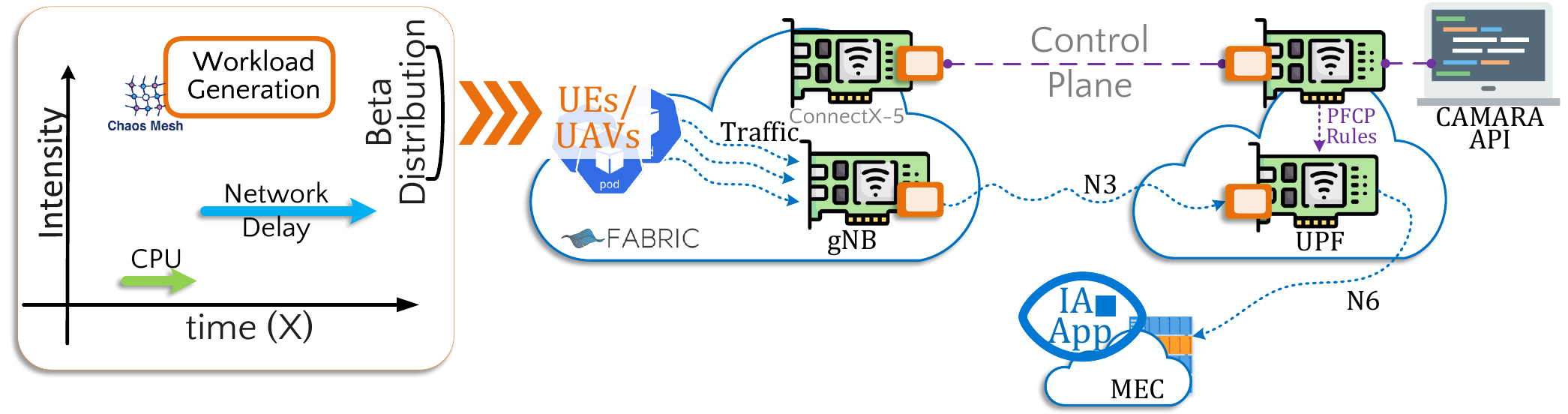}
    \caption{Our testbed.}
    \label{fig:testbed}
\end{figure}

We evaluate our approach\footnote{\scriptsize{Available at \url{https://github.com/romoreira/NetAI-APP-Inference_on_Demand}}.} on the FABRIC testbed~\cite{fabric-2019}. Figure~\ref{fig:testbed} depicts our testbed. Traffic is generated as \ac{GTP-U} slices, by a Kubernetes cluster with multiple pods acting as video source generation and forwarded through a \ac{BESS}-based \ac{UPF}~\cite{onf_aether}, which decapsulates the tunnels and extract the \ac{TLV} and process the tensor data accordingly, and forwards (N6 interface) toward \ac{MEC} or the data network.

\ac{UE} stress is generated (through ChaosMesh tool~\cite{chaosmesh}) independently per pod as a seeded event process: event starts follow exponential inter-arrivals with mean $35$~s over each $180$-sstage, while event durations are sampled uniformly from $10$ to $45$~s and may overlap. For each event, ChaosMesh applies \ac{CPU} stress with four workers and a load sampled from a bounded $\mathrm{Beta}(2,2)$ distribution over $[15,85]\%$, along with network delay and jitter sampled from bounded $\mathrm{Beta}(2,2)$ distributions over $[5,90]$~ms and $[1,20]$~ms, respectively.

\section{Results and Discussion}\label{sec:results}

\subsection{Dataset and Neural Network Training}\label{subsec:used_dataset}

%\begin{figure}[htbp]
%\centering
%\includegraphics[width=\columnwidth]{images/flame2_class_samples.pdf}
%\caption{Representative FLAME~2 RGB frames used in the binary fire-detection task.}
%\label{fig:flame2_samples}
%\end{figure}

We use FLAME~2 (\textit{Fire detection and modeLing: Aerial Multi-spectral imagE dataset}) as the visual workload for the split-inference service~\cite{swyw-6j78-22}. FLAME~2 is originally a \ac{UAV} video dataset, containing raw RGB/IR video pairs and labeled frame pairs derived from those videos; in this work, we use the 254p RGB frames and the binary \textit{Fire/NoFire} task.

The split model is derived from a ResNet18 backbone pre-trained on ImageNet, with the final fully connected layer replaced by a two-class classifier. Frames are resized to $224\times224$ and normalized with ImageNet statistics; training uses random horizontal flips, small rotations, and color jitter. To reduce temporal leakage between adjacent video frames, samples are grouped into temporal blocks of 1000 consecutive frames before the train/validation/test split. The baseline model is trained for 10 epochs with batch size 128, AdamW optimizer, learning rate $10^{-4}$, weight decay $10^{-4}$, cosine annealing, mixed precision on CUDA, and class-weighted cross-entropy.

\begin{figure}[htbp]
\centering
\subfloat[Learning curves of the baseline model.]{%
\includegraphics[width=0.48\linewidth]{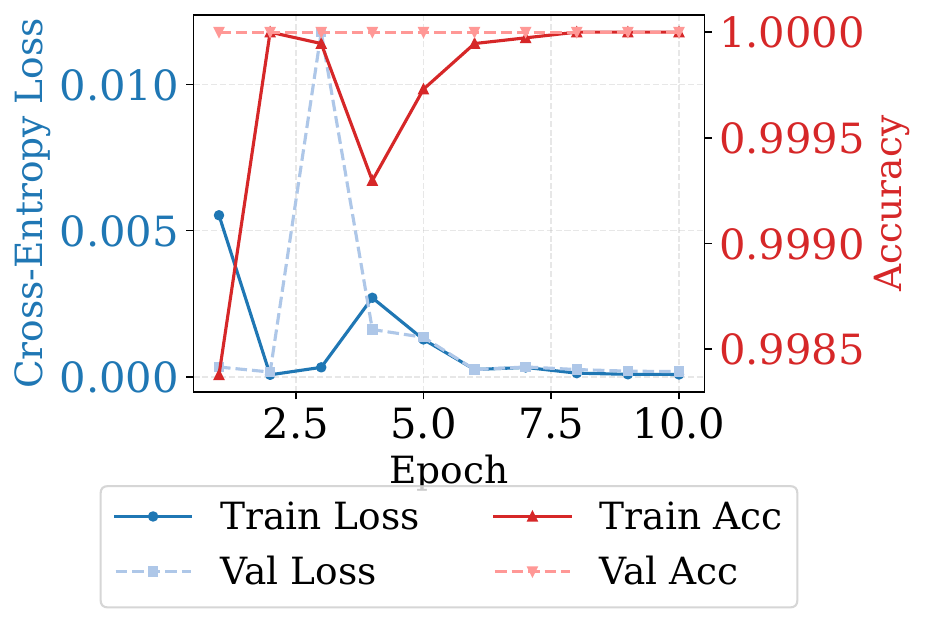}%
\label{fig:learning_curves}%
}%
\hfill%
\subfloat[Confusion matrix on the test split.]{%
\includegraphics[width=0.43\linewidth]{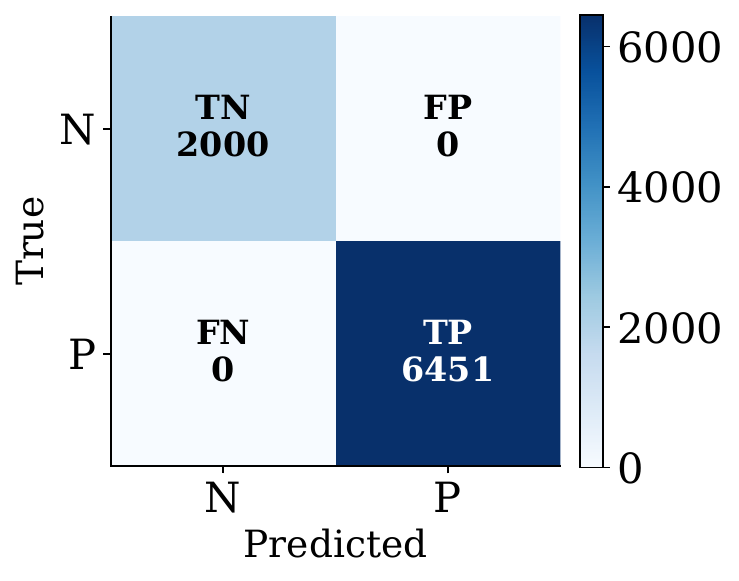}%
\label{fig:confusion_matrix}%
}
\caption{Baseline model validation on FLAME~2: (a) training and validation dynamics over epochs and (b) the resulting test-set confusion matrix.}
\label{fig:baseline_results}
\end{figure}

Figure~\ref{fig:baseline_results} reports the convergence behavior and the final confusion matrix of the trained baseline. This model is not the contribution of the paper; it provides a validated partitionable visual workload whose intermediate activations are used by the static and adaptive split-inference experiments.

\subsection{Inference-on-Demand Split Adaptation}\label{subsec:iod_results}

We carried out experiments in order to evaluate our approach. We aim to answer whether \ac{IoD} can operate as a dynamic split-selection mechanism under changing user-plane load. Our results show whether \ac{IoD} can recover a useful operating region without selecting $k$ a priori, react to increasing contention, and keep the \ac{UPF} load below bytes-heavy static configurations.

\begin{figure}[htbp]
\centering
\includegraphics[width=\columnwidth]{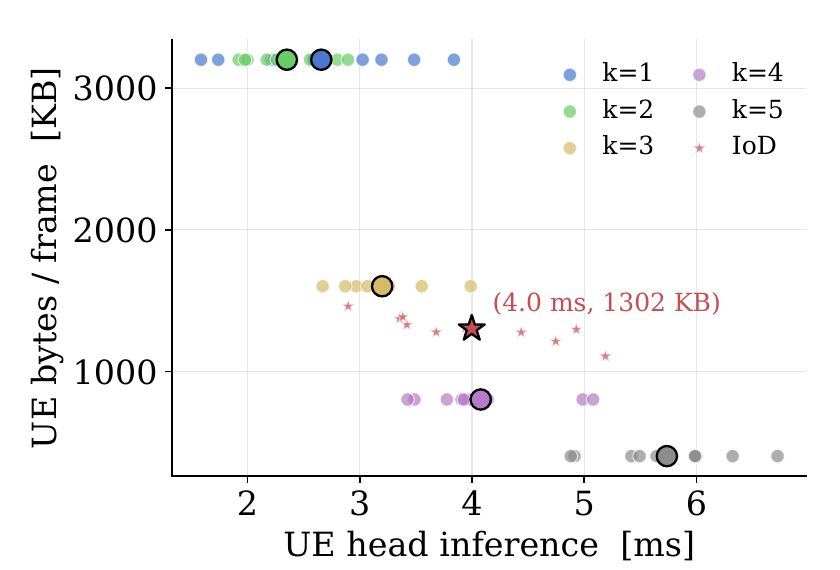}
\caption{Compute--bytes operating frontier across fixed split points and IoD. IoD operates near the middle of the frontier without a priori selection of $k$.}
\label{fig:pareto}
\end{figure}

Figure~\ref{fig:pareto} shows the compute--bytes operating frontier. The static baselines follow the expected trade-off: shallow splits reduce \ac{UE}-side head inference time but transmit more bytes, whereas deeper splits reduce transmitted bytes at the cost of more \ac{UE}-side computation. The \ac{IoD} operating point lies near the middle of this frontier, at approximately $4.0$~ms of UE head inference and $1302$~KB per frame. This is close to the region represented by $k{=}3$, but it is reached without configuring a fixed split in advance. This result shows that the adaptive policy recovers a balanced operating point dynamically.

\begin{figure}[htbp]
\centering
\includegraphics[width=\columnwidth]{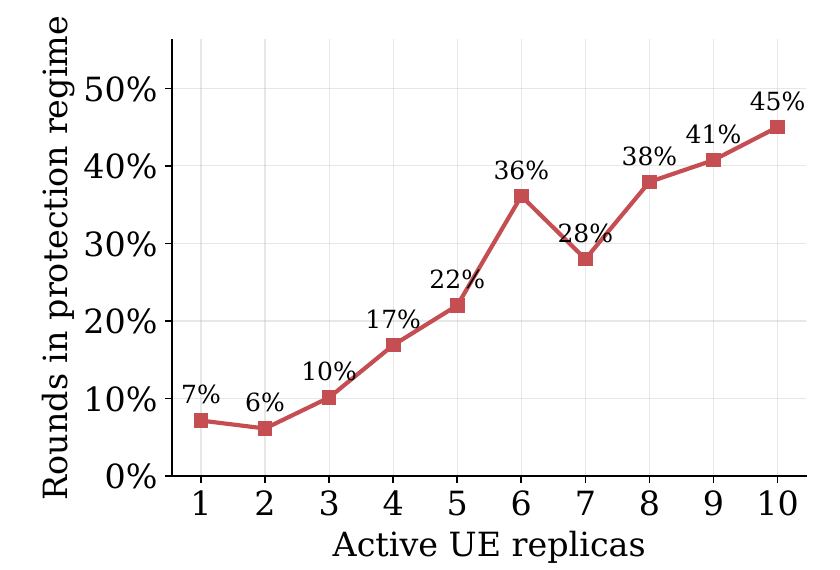}
\caption{Fraction of IoD rounds in the network-protection regime as the number of active UE replicas increases.}
\label{fig:adapt}
\end{figure}

\begin{figure}[htbp]
\centering
\includegraphics[width=\columnwidth]{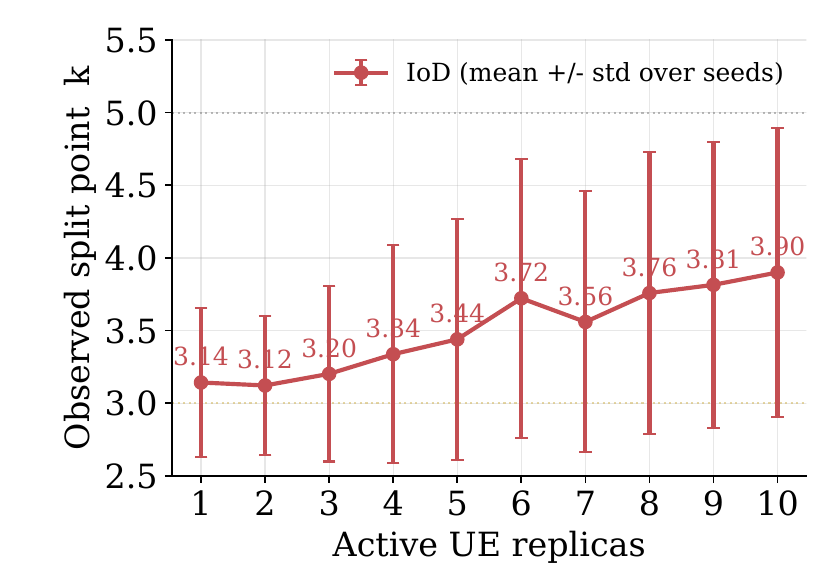}
\caption{Observed IoD split point as a function of active UE replicas. Error bars show variation across seeds.}
\label{fig:ktraj}
\end{figure}

Figures~\ref{fig:adapt} and~\ref{fig:ktraj} show how this operating point changes as the number of active \ac{UE} replicas increases. In Figure~\ref{fig:adapt}, the fraction of rounds in the protection regime rises from $7\%$ with one active replica to $45\%$ with ten active replicas. This indicates that the controller uses the protection branch more often as contention increases. The trend is not
strictly monotonic, which is expected because each seed produces its own scale and chaos sequence, but the overall direction is clear. Figure~\ref{fig:ktraj} confirms the same behavior through the observed split point: the mean value of $k$ increases from $3.14$ to $3.90$ across the same scale range. The wider variation at higher replica counts indicates that \ac{IoD} does not simply replace one static split with another. It alternates between regimes at runtime according to the measured condition of each round.

\begin{figure}[htbp]
\centering
\includegraphics[width=\columnwidth]{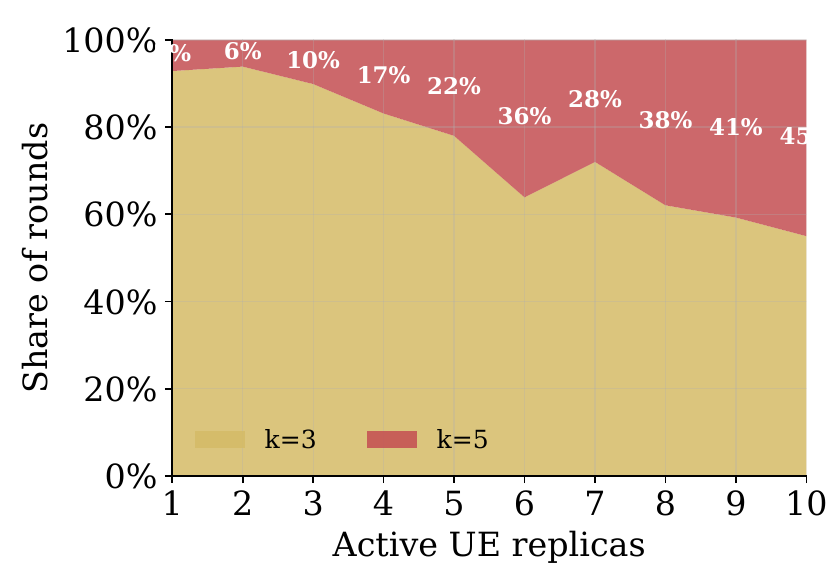}
\caption{Share of IoD rounds assigned to the nominal and network-protection regimes.}
\label{fig:regime}
\end{figure}

Figure~\ref{fig:regime} clarifies which controller branches were exercised by the campaign. Two regimes dominate the execution: the nominal regime, represented by $k{=}3$, and the network-protection regime, represented by $k{=}5$. At one replica, $93\%$ of the rounds remain in the nominal regime, whereas at ten replicas this share falls to $55\%$. Conversely, the protection share increases from $7\%$ to $45\%$. The other designed branches were not activated because the workload did not create \ac{UE}-side compute pressure: head inference remained far below the configured $80$~ms threshold. Therefore, this campaign demonstrates the network-protection behavior of \ac{IoD}, while the compute-protection branches should be treated as part of the controller design rather than as empirically exercised regimes in this experiment.

\begin{figure}[htbp]
\centering
\includegraphics[width=\columnwidth]{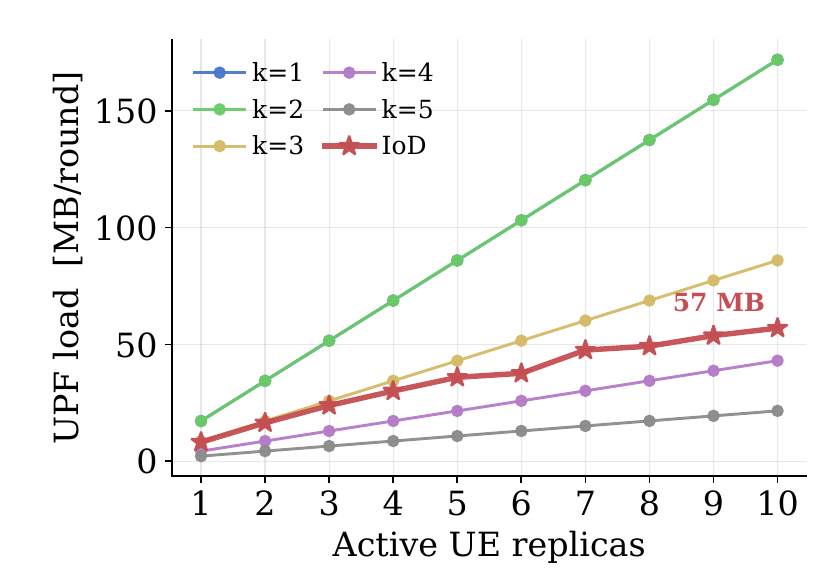}
\caption{UPF-counted load under increasing numbers of active UE replicas.}
\label{fig:upfload}
\end{figure}

Figure~\ref{fig:upfload} reports the \ac{UPF}-counted load as the number of active \ac{UE} replicas changes. The static baselines grow almost linearly with scale. At ten replicas, the shallowest split reaches approximately $171.84$~MB per round, whereas $k{=}3$ reaches $85.95$~MB and $k{=}5$ reaches $21.53$~MB. \ac{IoD} reaches $56.97$~MB per round at the same scale, about one third of the load imposed by $k{=}1$. This confirms that \ac{IoD} substantially reduces user-plane load compared with bytes-heavy split points. The shape of the \ac{IoD} curve is also informative: between five and six replicas, the load increases only from $35.89$~MB to $37.62$~MB, which coincides with the controller using the protection regime more often. Thus, the reduction is not only visible in the average operating point; it also appears as a scale-dependent response in the \ac{UPF} load curve.

\begin{figure}[htbp]
\centering
\includegraphics[width=\columnwidth]{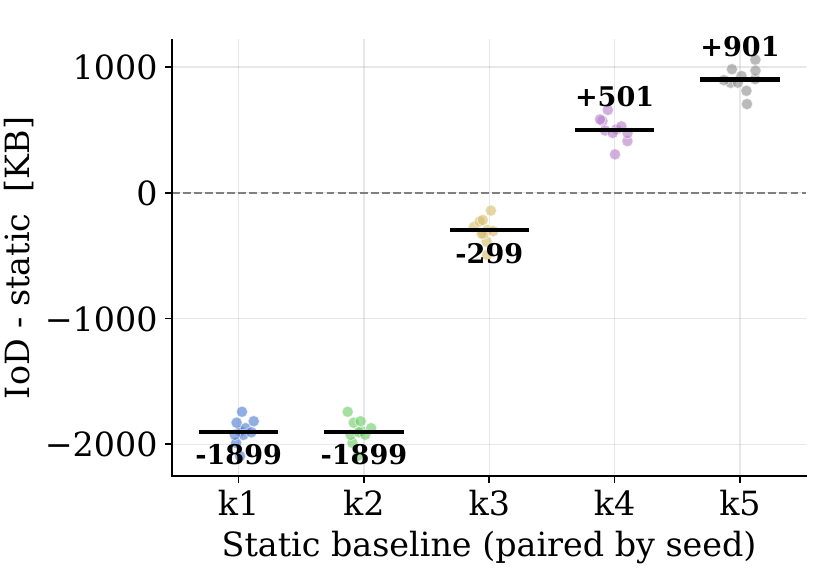}
\caption{Per-seed paired difference between IoD and each static baseline in UE bytes per frame. Negative values indicate that IoD transmits fewer bytes than the corresponding static split.}
\label{fig:paired}
\end{figure}

Finally, Figure~\ref{fig:paired} compares \ac{IoD} with each static baseline using paired seeds. Negative values indicate that \ac{IoD} transmits fewer bytes per frame than the corresponding static split. The result is consistent across seeds: \ac{IoD} reduces transmitted bytes by approximately $1899$~KB per frame relative to $k{=}1$ and $k{=}2$, and by approximately $299$~KB relative to $k{=}3$. As expected, \ac{IoD} transmits more bytes than the more compute-heavy configurations, with average differences of approximately $501$~KB relative to $k{=}4$ and $901$~KB relative to $k{=}5$. This paired view reinforces the intended reading of the method: \ac{IoD} is not a universal minimizer of bytes or computation. It is a dynamic compromise and a novel mobile network data plane service that avoids the high communication cost of shallow splits, stays near a balanced operating region, and increases protection when the network becomes more pressured.

Overall, the results support the feasibility of \ac{IoD} as an online adaptation mechanism for split inference over a mobile user-plane path. The method recovers a balanced operating region without prior split selection, reacts to increasing contention by shifting toward protection, and keeps \ac{UPF} load below bytes-heavy static baselines.

\section{Concluding Remarks}\label{sec:concluding_remarks}

This work addressed the gap between edge inference as an application-side offloading mechanism and edge inference as a programmable telco service. The results show that the \ac{IoD} can adapt the split point at runtime, keep tensor transport observable in the \ac{5G} user plane, and reduce the pressure imposed by bytes-heavy static configurations. This finding suggests that mobile networks can support \ac{AI} execution not only by providing connectivity but also by exposing service intent and carrying inference context along the data path. Future work will extend \ac{IoD} toward tighter CAMARA \ac{API} integration, policy-driven split selection, and larger multi-tenant \ac{MEC} deployments. Overall, the \ac{IoD} reframes split inference as a network-native service primitive for adaptive \ac{AI} over a programmable mobile infrastructure.

\section*{Acknowledgment}

The authors thank FAPEMIG (Grant \#APQ00923-24), FAPESP MCTIC/CGI Research project 2018/23097-3 - SFI2 - Slicing Future Internet Infrastructures, Coordenação de Aperfeiçoamento de Pessoal de Nível Superior - Brasil (CAPES) - Finance Code 001, PROPP/UFU, and Fundação para a Ciência e Tecnologia (FCT) within the R\&D Unit Project Scope UID/00319/2025 - Centro ALGORITMI (ALGORITMI/UM) for supporting this work. The authors also acknowledge IMAGINE-B5G (EU Horizon Europe, Grant No. 101096452) for the experimental infrastructure used in this work.



\bibliographystyle{IEEEtran}
\bibliography{example}

@misc{camaraproject,
  author = {{Linux Foundation} and {GSMA}},
  title  = {{CAMARA}: The {Telco Global API Alliance}},
  year   = {2024},
  url    = {https://camaraproject.org/},
  note   = {Acesso em: \today}
}

@misc{onf_aether,
  author       = {{Open Networking Foundation (ONF)}},
  title        = {Aether: Open Source 5G Connected Edge Platform},
  howpublished = {\url{https://opennetworking.org/aether/}},
  year         = {2021},
  note         = {Acessado em: 18 de junho de 2026}
}

@article{fabric-2019,
              title={{FABRIC: A national-scale programmable experimental network infrastructure}},
              author={Baldin, Ilya and Nikolich, Anita and Griffioen, James and Monga, Indermohan Inder S and Wang, Kuang-Ching and Lehman, Tom and Ruth, Paul},
              journal={IEEE Internet Computing},
              volume={23},
              year={2019},
              number={6},
              pages={38--47},
              publisher={IEEE}
            }

@misc{chaosmesh,
  author       = {{Chaos Mesh Authors}},
  title        = {{Chaos Mesh: A Powerful Chaos Engineering Platform for Kubernetes}},
  howpublished = {\url{https://chaos-mesh.org/}},
  year         = {2020}, 
  note         = {GitHub repository: \url{https://github.com/chaos-mesh/chaos-mesh}}
}

@data{swyw-6j78-22,
doi = {10.21227/swyw-6j78},
url = {https://dx.doi.org/10.21227/swyw-6j78},
author = {Bryce Hopkins and Leo O'Neill and Fatemeh Afghah and Abolfazl Razi and Eric Rowell and Adam Watts and Peter Fule and Janice Coen},
publisher = {IEEE Dataport},
title = {FLAME 2: Fire detection and modeLing: Aerial Multi-spectral imagE dataset},
year = {2022} }

@INPROCEEDINGS{paper1, author={Djuhera, Aladin and Koch, Fernando and Binotto, Alecio}, booktitle={2026 International Conference on Computing, Networking and Communications (ICNC)}, title={Joint Partitioning and Placement of Foundation Models for Real-Time Edge AI}, year={2026}, volume={}, number={}, pages={459-464}, doi={10.1109/ICNC68183.2026.11416947}}

@ARTICLE{paper2,
  author={Shah, Syed Danial Ali and Ali, Farman and AlZubi, Ahmad Ali and Kwak, Daehan},
  journal={IEEE Consumer Electronics Magazine}, 
  title={Collaborative Intelligence: Orchestrating Ultra-Low Latency XR Experiences in 6G Consumer Networks}, 
  year={2026},
  volume={},
  number={},
  pages={1-11},
  doi={10.1109/MCE.2026.3698602}}

@article{paper3,
   title={Intelligent Orchestration of Distributed Large Foundation Model Inference at the Edge},
   ISSN={2972-4619},
   url={http://dx.doi.org/10.37256/cnc.3220256807},
   DOI={10.37256/cnc.3220256807},
   journal={Computer Networks and Communications},
   publisher={Universal Wiser Publisher Pte. Ltd},
   author={Koch, Fernando and Djuhera, Aladin and Binotto, Alecio},
   year={2025},
   month=Sept, pages={111–128} }

@ARTICLE{paper4,
  author={Zhang, Guanlei and Zhang, Qiyang and Feng, Lei and Zhou, Fanqin and Donta, Praveen Kumar and Dustdar, Schahram},
  journal={IEEE Transactions on Mobile Computing}, 
  title={Task-Aware Collaborative Inference and Fine-Grained DNN Partitioning in MEC Networks}, 
  year={2026},
  volume={25},
  number={6},
  pages={8911-8927},
  doi={10.1109/TMC.2025.3650680}}

@ARTICLE{paper5,
  author={Tran, Minh-Ngoc and Nguyen Trung, Kiem and Kim, Younghan},
  journal={IEEE Access}, 
  title={Design of 5G Architecture Enhancements for Supporting Edge Split Computing Service Pipeline}, 
  year={2025},
  volume={13},
  number={},
  pages={191515-191530},
  doi={10.1109/ACCESS.2025.3630182}}

@ARTICLE{paper6,
  author={Lim, Jeong-A and Lee, Joohyun and Kwak, Jeongho and Kim, Yeongjin},
  journal={IEEE Transactions on Services Computing}, 
  title={Cutting-Edge Inference: Dynamic DNN Model Partitioning and Resource Scaling for Mobile AI}, 
  year={2024},
  volume={17},
  number={6},
  pages={3300-3316},
  doi={10.1109/TSC.2024.3466848}}

@ARTICLE{paper7,
  author={Fang, Chuxing and Xu, Changqiao and Li, Zitong and Xiao, Han and Yang, Shujie and Zhong, Lujie and Muntean, Gabriel-Miro},
  journal={IEEE Transactions on Mobile Computing}, 
  title={Non-intrusive Handover Strategy Optimization for Model-Partitioned DNN Inference in Satellite Edge Computing}, 
  year={2026},
  volume={},
  number={},
  pages={1-17},
  doi={10.1109/TMC.2026.3674471}}

@ARTICLE{paper8,
  author={He, Jia and Wu, Huanzhuo and Xiao, Xun and Bassoli, Riccardo and Fitzek, Frank H. P.},
  journal={IEEE Wireless Communications}, 
  title={Functional Split of In-Network Deep Learning for 6G: A Feasibility Study}, 
  year={2022},
  volume={29},
  number={5},
  pages={36-42},
  doi={10.1109/MWC.003.2200060}}

@article{paper9,
title = {Mobility-Aware Predictive Split Federated Learning for 6G vehicular networks with ultra-low latency guarantees},
journal = {Computer Networks},
volume = {270},
pages = {111553},
year = {2025},
issn = {1389-1286},
doi = {https://doi.org/10.1016/j.comnet.2025.111553},
url = {https://www.sciencedirect.com/science/article/pii/S1389128625005201},
author = {Iftikhar Rasheed and Hala Mostafa}
}

@INPROCEEDINGS{paper10,
  author={Lim, Jeong-A and Kim, Yeongjin},
  booktitle={2022 19th Annual IEEE International Conference on Sensing, Communication, and Networking (SECON)}, 
  title={Real-time DNN Model Partitioning for QoE Enhancement in Mobile Vision Applications}, 
  year={2022},
  volume={},
  number={},
  pages={407-415},
  doi={10.1109/SECON55815.2022.9918600}}

@article{paper11,
title = {MAE: Collaborative inference acceleration with efficient DNN partitioning and resource allocation in resource-constrained edge computing},
journal = {Computer Networks},
volume = {278},
pages = {112073},
year = {2026},
issn = {1389-1286},
doi = {https://doi.org/10.1016/j.comnet.2026.112073},
url = {https://www.sciencedirect.com/science/article/pii/S138912862600085X},
author = {Juan Fang and Yaxin An and Yaqi Liu and Ziyi Teng and Xiaoning Zhai and Heng Tang and Huijie Chen}
}

@ARTICLE{paper12,
  author={Tan, Lin and Guo, Kehan and Tan, Zhiya and Guo, Songtao and Kuang, Zhufang and Zhao, Jun and Niyato, Dusit},
  journal={IEEE Transactions on Mobile Computing}, 
  title={DA-ERL: Demand-Aware Partitioned Collaborative Inference for On-Device Models}, 
  year={2026},
  volume={},
  number={},
  pages={1-17},
  doi={10.1109/TMC.2026.3668325}}

@INPROCEEDINGS{paper13,
  author={Tan, Lin and Zhou, Pengzhan and Guo, Songtao and Zhao, Jun and Kuang, Zhufang and Qiao, Dewen and Yang, Lu},
  booktitle={2025 IEEE 45th International Conference on Distributed Computing Systems (ICDCS)}, 
  title={Partitioned Collaborative Inference for On-Device Models via Evolutionary Reinforcement Learning}, 
  year={2025},
  volume={},
  number={},
  pages={714-724},
  doi={10.1109/ICDCS63083.2025.00075}}

@INPROCEEDINGS{paper14,
  author={Guo, Kai and Xiao, Yilin and Lu, Xiaozhen and Xiao, Liang},
  booktitle={2025 IEEE Wireless Communications and Networking Conference (WCNC)}, 
  title={Learning-Based Low-Latency Collaborative Inference for Multi-Branch Models in D2D-Assisted MEC}, 
  year={2025},
  volume={},
  number={},
  pages={01-06},
  doi={10.1109/WCNC61545.2025.10978329}}

@misc{gsma_opengateway_alt,
  author       = {{GSMA}},
  title        = {GSMA Open Gateway Initiative},
  year         = {2024},
  url          = {https://www.gsma.com/opengateway},
  note         = {Acessado em: 29 jun. 2026}
}

@ARTICLE{rodrigo2026,
  author={Moreira, Rodrigo and Moreira, Larissa Ferreira Rodrigues and Silva, Flávio de Oliveira},
  journal={IEEE Access}, 
  title={Unleashing AI-Empowered Slices on Mobile Networks for Natively Cognitive Service Delivery}, 
  year={2025},
  volume={13},
  number={},
  pages={147757-147771},
  doi={10.1109/ACCESS.2025.3599356}}

@inproceedings{Ordonez-Lucena2022,
author = {Ordonez-Lucena, Jose and Dsouza, Felix},
title = {Pathways towards network-as-a-service: the CAMARA project},
year = {2022},
isbn = {9781450393959},
publisher = {Association for Computing Machinery},
address = {New York, NY, USA},
url = {https://doi.org/10.1145/3538401.3546825},
doi = {10.1145/3538401.3546825},
booktitle = {Proceedings of the ACM SIGCOMM Workshop on Network-Application Integration},
pages = {53–59},
numpages = {7},
location = {Amsterdam, Netherlands},
series = {NAI '22}
}

@ARTICLE{Tang2026,
  author={Tang, Xun and Li, Qing and Peng, Junkun and Tan, Yuanzheng and Yuan, Zhenhui and Song, Yuhong and Muntean, Gabriel-Miro and Jiang, Yong},
  journal={IEEE Communications Surveys \& Tutorials}, 
  title={Low-Latency Video Streaming: Applications, Challenges, and Trends}, 
  year={2026},
  volume={28},
  number={},
  pages={6384-6414},
  doi={10.1109/COMST.2026.3689266}}

@ARTICLE{Niu2026,
  author={Niu, Xupeng and Yuan, Weijie and Tan, Long and Su, Nanchi and Zhang, Yunyang and Quek, Tony Q. S.},
  journal={IEEE Communications Magazine}, 
  title={Beyond Connectivity: How Large AI Models Unlock Semantic 6G Networks?}, 
  year={2026},
  volume={64},
  number={4},
  pages={106-112},
  doi={10.1109/MCOM.001.2500456}}

@INPROCEEDINGS{Lessa2025,
  author={Fatur Lessa, João Vicente and Marques, Jonatas Adilson and Parizotto, Ricardo and Gaspary, Luciano Paschoal},
  booktitle={NOMS 2025-2025 IEEE Network Operations and Management Symposium}, 
  title={In-Network Inference with Neuralp4: Auto-Generating Full-Fledged NN Models for PDPs}, 
  year={2025},
  volume={},
  number={},
  pages={1-10},
  doi={10.1109/NOMS57970.2025.11073576}}

@ARTICLE{Kim2026,
  author={Kim, Heewon and Yoon, Seongyeon and Bae, Chanbin and Lee, Sanghoon and Pack, Sangheon},
  journal={IEEE Transactions on Networking}, 
  title={TINIEE: Traffic-Aware Adaptive In-Network Intelligence via Early-Exit Strategy}, 
  year={2026},
  volume={34},
  number={},
  pages={5037-5052},
  doi={10.1109/TON.2026.3689743}}

@article{Lin2025,
title = {NDIF: A distributed framework for efficient in-network neural network inference},
journal = {Computers \& Security},
volume = {157},
pages = {104593},
year = {2025},
issn = {0167-4048},
doi = {https://doi.org/10.1016/j.cose.2025.104593},
url = {https://www.sciencedirect.com/science/article/pii/S0167404825002822},
author = {Shengrui Lin and Shaowei Xu and Binjie He and Hongyan Liu and Dezhang Kong and Xiang Chen and Dong Zhang and Chunming Wu and Ming Li and Xuan Liu and Yuqin Wu and Muhammad Khurram Khan}
}

@ARTICLE{Zhang2026,
  author={Zhang, Guanlei and Zhang, Qiyang and Feng, Lei and Zhou, Fanqin and Donta, Praveen Kumar and Dustdar, Schahram},
  journal={IEEE Transactions on Mobile Computing}, 
  title={Task-Aware Collaborative Inference and Fine-Grained DNN Partitioning in MEC Networks}, 
  year={2026},
  volume={25},
  number={6},
  pages={8911-8927},
  doi={10.1109/TMC.2025.3650680}}

@ARTICLE{Liu2026,
  author={Liu, Jing and Du, Yao and Yang, Kun and Wu, Jiaqi and Wang, Yan and Hu, Xiping and Wang, Zehua and Liu, Yang and Sun, Peng and Boukerche, Azzedine and Leung, Victor C. M.},
  journal={IEEE Communications Surveys \& Tutorials}, 
  title={Edge-Cloud Collaborative Computing on Distributed Intelligence and Model Optimization: A Survey}, 
  year={2026},
  volume={28},
  number={},
  pages={5049-5080},
  doi={10.1109/COMST.2026.3669216}}

\end{document}